\documentclass[conference]{IEEEtran}
\IEEEoverridecommandlockouts
\usepackage{cite}
\usepackage{authblk}
\usepackage{varwidth}
\usepackage{graphicx}
\usepackage{xspace}
\usepackage{xcolor}
\usepackage{comment}
\usepackage{geometry}
\usepackage{amsmath}
\usepackage{hyperref}
\usepackage{multirow}
\usepackage{multicol}
\usepackage{color}
\usepackage{adjustbox}
\usepackage{xcolor}
\usepackage{colortbl}
\usepackage{booktabs}
\usepackage{flushend}
\usepackage{booktabs}
\usepackage{tabularx}
\usepackage{caption}
\usepackage{booktabs}
\renewcommand{\IEEEtitletopspaceextra}{\dimexpr-\headheight-\headsep+18pt\relax}

\usepackage{cite}
\usepackage{amsmath,amssymb,amsfonts}
\usepackage{algorithmic}
\usepackage{graphicx}
\usepackage{textcomp}
\usepackage{xcolor}
\def\BibTeX{{\rm B\kern-.05em{\sc i\kern-.025em b}\kern-.08em
    T\kern-.1667em\lower.7ex\hbox{E}\kern-.125emX}}

\begin{document}

\title{Improving the Factual Accuracy of Abstractive Clinical Text Summarization using Multi-Objective Optimization\\



\thanks{Research reported in this publication was supported by NCCIH of the National Institutes of Health under award number 5R01AT010413S1.}
}

\author[1]{Amanuel Alambo\thanks{Corresponding author: Amanuel Alambo (\textcolor{blue}{alambo.2@wright.edu})}}

\author[1]{Tanvi Banerjee}
\author[1]{Krishnaprasad Thirunarayan}
\author[2]{Mia Cajita}

\affil[1]{Wright State University} 
\affil[2]{University of Illinois, Chicago}

\maketitle

\begin{abstract}
While there has been recent progress in abstractive summarization as applied to different domains including news articles, scientific articles, and blog posts, the application of these techniques to clinical text summarization has been limited. This is primarily due to the lack of large-scale training data and the messy/unstructured nature of clinical notes as opposed to other domains where massive training data come in structured or semi-structured form. Further, one of the least explored and critical components of clinical text summarization is factual accuracy of clinical summaries. This is specifically crucial in the healthcare domain, cardiology in particular,  where an accurate summary generation that preserves the facts in the source notes is critical to the well-being of a patient. In this study, we propose a framework for improving the factual accuracy of abstractive summarization of clinical text using knowledge-guided multi-objective optimization. We propose to jointly optimize three cost functions in our proposed architecture during training: \emph{generative loss, entity loss and knowledge loss} and evaluate the proposed architecture on 1) clinical notes of patients with heart failure (HF), which we collect for this study; and 2) two benchmark datasets, Indiana University Chest X-ray collection (IU X-Ray), and MIMIC-CXR, that are publicly available. We experiment with three transformer encoder-decoder architectures and demonstrate that optimizing different loss functions leads to improved performance in terms of entity-level factual accuracy.
\end{abstract}


\section{Introduction}

Recent advances in transformer-based models \cite{vaswani2017attention} have led to progress in abstractive summarization of news articles, scientific articles, and social media data. However, these models have not been well investigated in the healthcare domain where automated clinical summary generation \cite{pivovarov2015automated} for a set of findings in clinical notes is helpful to clinicians in saving their time and improving clinical decision making. One of the clinical practices by medical professionals entails the task of recording \emph{findings} of diagnosis, treatment or procedures followed by summarizing the findings into a form called \emph{impression}. Inspired by recent efforts in modeling findings-to-impression as summarization \cite{zhang2018learning, zhang2019optimizing, macavaney2019ontology}, we propose to automate this process of writing an impression for findings to assist clinicians with their practice, making the clinical workflow more efficient. Specifically, we attempt to accomplish this using an abstractive approach to summarizing findings into an impression. Further, clinicians use their commonsense understanding and their knowledge of the domain while producing an impression in addition to what is explicitly stated in the findings. As such, an impression has to be factually correct with respect to the findings. This is particularly critical in the healthcare setting where a misinterpreted impression could prove fatal and should be avoided at all costs to deliver quality health to patients. This issue is further exacerbated in the sub-domain of \emph{Heart Failure} (HF) where reliable diagnosis is challenging and the cost of inaccuracy can be enormous.  To investigate these issues, we utilize clinical notes of 1200 patients with HF from the University of Illinois Hospital \& Health Sciences System (UI Health) for our study.  Figure-1 illustrates what a typical clinical note (a record) for a patient with HF in our cohort looks like.

\begin{figure}[h]
  \centering
  \includegraphics[height=11cm, width=0.5\textwidth]{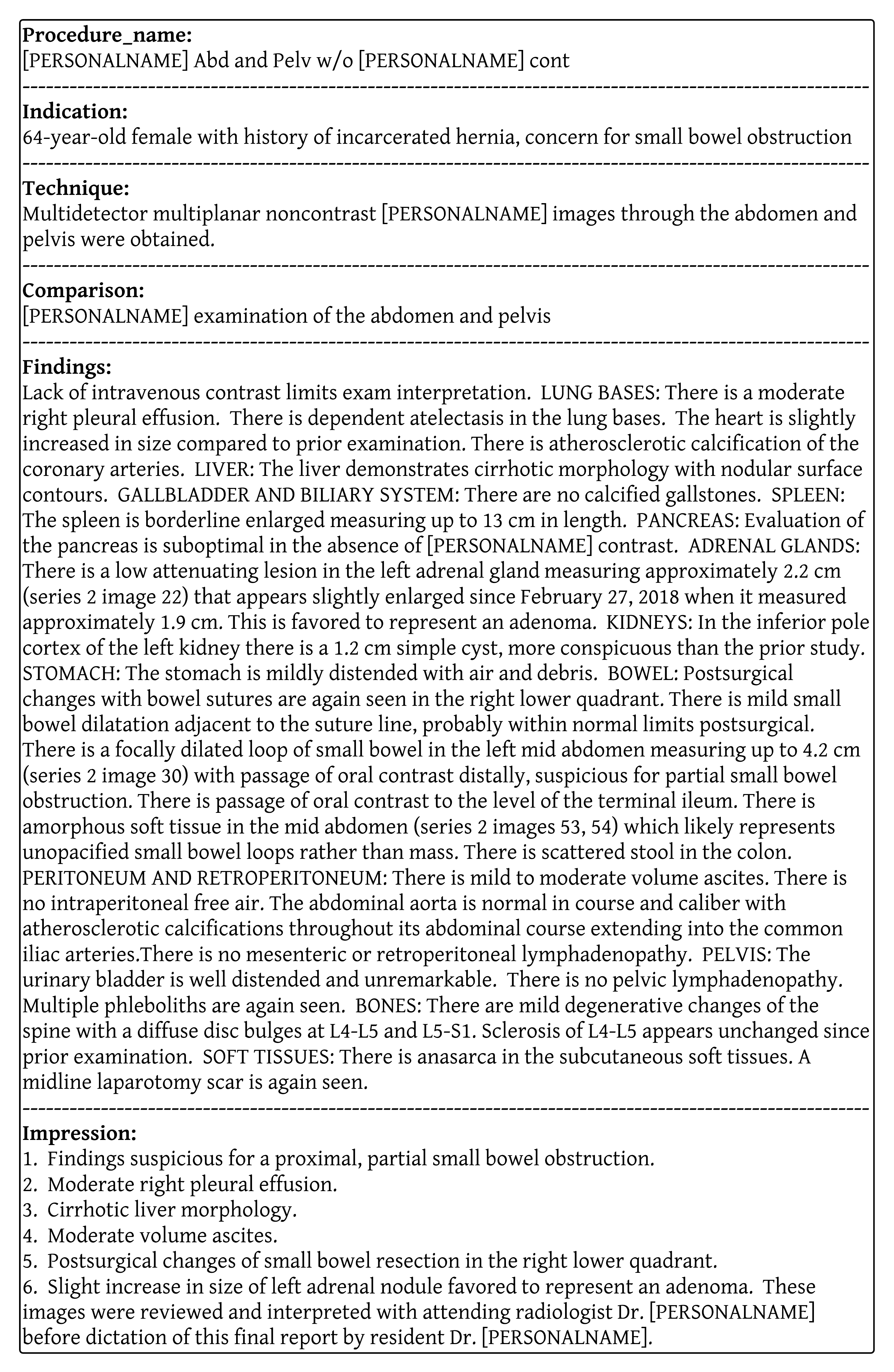}
  \caption{Example de-identified clinical record for the heart failure data collected through the Center for Clinical and Translational Science, University of Illinois, Chicago.}
\vspace{-4mm}
\end{figure}

In this paper, we model and automate this clinician’s impression (summary) writing process using a two-stage approach: 1) clinical knowledge retrieval from domain-specific knowledge sources using named entities; and 2) joint training of end-to-end transformer encoder-decoder models using multi-objective optimization. We evaluate our proposed framework on two benchmark datasets and a dataset we prepare for this task (heart-failure - HF) and demonstrate that we achieve significantly better results over baseline model training settings of \emph{ findings-to-impression} on factual accuracy metrics. We use three SOTA pre-trained transformer-encoder-decoder networks and fine-tune them on our datasets using different training objectives and report the results of the fine-tuning for each dataset. Our experimental data for HF consists of 6182 patient records where each record comprises \textit{Procedure Type, Techniques, Indication, Findings} and \textit{Impression} tuples of clinical notes. 

The main contributions of this study are: 1) the introduction of an approach for clinical named entity-aware knowledge retrieval from medical knowledge sources; 2) a novel training technique for abstractive summarization of clinical text using multi-objective optimization; and 3) new experiments using the state-of-the-art transformer encoder-decoder networks as backbone models to demonstrate that optimizing knowledge-driven cost functions in addition to the generative cost function during training boosts model performance on factual accuracy metrics.

\section{Related Work}

The birth of Transformer encoder-decoder models \cite{vaswani2017attention, raffel2019exploring, lewis2019bart, zhang2020pegasus, qi2020prophetnet} has led to significant advances in abstractive summarization in the domains of news articles \cite{hermann2015teaching, rush2015neural, see2017get}, and scientific articles \cite{cohan2018discourse, cachola2020tldr, yasunaga2019scisummnet}. Nevertheless, their application to the summarization of clinical notes has not been adequately explored. \cite{macavaney2019ontology} proposed a model based on Pointer-Generator-Networks \cite{see2017get} for abstractive summarization of radiology reports by linking entities in a clinical note to domain-specific ontology from UMLS \cite{bodenreider2004unified} and RadLex \cite{langlotz2006radlex}. They use pairings of findings and impression for the abstractive summarization task where findings form the input sequences and impressions form the target summaries for training. \cite{sotudeh2020attend} propose a two-stage model consisting of a content selector and abstractive summarizer for clinical abstractive summarization. The content selector identifies ontological terms from the findings using a medical ontology (RadLex) and the summarizer is trained to generate summaries (impressions). They use Bi-LSTMs to encode findings and use LSTMs to encode the ontological terms followed by an LSTM-based decoder to generate a summary. \cite{liang2019novel} built a model for extractive summarization of clinical notes of patients with diabetes and hypertension to generate disease-specific summaries. They framed the summary generation problem as a sentence classification problem and experimented on a dataset consisting of 3,453 clinical notes collected for 762 patients. \cite{weng2020clinical} proposed a model comprised of syntax-based negation detection and semantic clinical concept recognition for extractive summarization of clinical text. They conducted their experiments on the MIMIC-III \cite{johnson2016mimic} dataset. While the aforementioned approaches employ different techniques for clinical text summarization, we show experimentally that our proposed knowledge-aware Multi-Objective Optimization (MOO) improves the factual accuracy of the generated summaries when compared to strong state-of-the-art transformer-based abstractive summarization models.

\vspace{-1mm}
\section{Data Preparation}
Out of the total of 15183 de-identified procedure notes spanning a period of over 4  years (5/2016 - 8/2020) collected from patients with HF admitted to UI Health, we filter the ones with no Findings or Impressions since our task is to generate an impression for a finding. Thus, the findings play the role of input text to be summarized and the impression serves as the ground truth summary.  After pre-processing the data, we have 6182 notes consisting of findings-to-impression pairings along with other metadata. In addition to our Heart Failure data, we evaluate the proposed approach on two benchmark datasets on radiology reports from the Indiana Network for Patient Care \cite{demner2016preparing} and 50000 randomly selected chest x-ray reports from the MIMIC-III-CXR dataset \cite{johnson2019mimic}.

\section{Proposed Approach}
\subsection{Clinical Text Named Entity and Knowledge Extraction}

We use an off-the-shelf Stanza package from Stanford for clinical named entity recognition (NER) \cite{qi2020stanza}. Specifically, the Stanza model we use is the one trained on the i2b2 clinical text dataset. The knowledge bases to query for facts using the named entities are composed of UMLS, SNOMED-CT, and ICD-10. After named entities are extracted using Stanza from a finding, our next task is to query for facts pertaining to the named entities as they appear in domain-specific knowledge bases. For each named entity identified from a finding, we perform full-text lexical query of the KBs and return the top-k facts where we set the value of $K$ to 5 \cite{an2021retrievalsum}.

\subsection{Model Training using Multi-Objective Optimization}
We experiment with three state-of-the-art transformer-based models pretrained using different self-supervised objectives. We propose to train these models using a loss function that optimizes summary generation, named entity chain generation, and fact generation where our task is not only to auto-regressively generate the target summary, but also to generate the named entities in the impression and to generate the facts associated with the named entities in the impression. Figure-2 shows the proposed end-to-end architecture where three networks, whose parameters are shared are jointly trained using the loss functions stated in Equation-1.

\begin{figure}[h]
  \centering
  \includegraphics[width=0.5\textwidth]{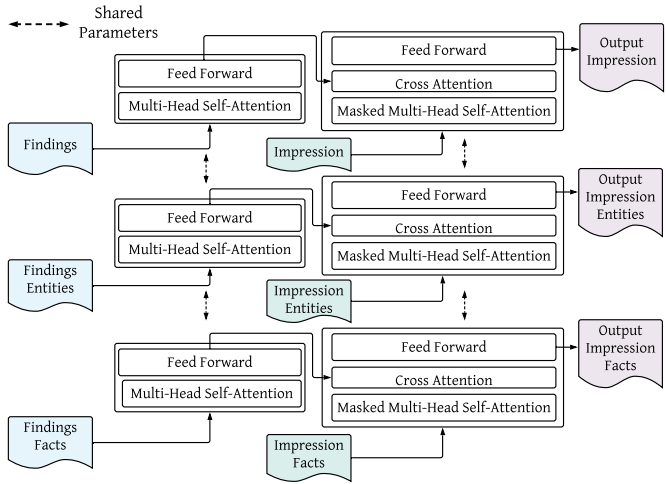}
  \caption{The proposed training architecture. }
\vspace{-3mm}
\end{figure}

We optimize the total aggregate loss function during the training phase for the proposed model in use. We use Bayesian optimization \cite{snoek2012practical} to search for the best combination of \emph{generative} and \emph{regularization} hyperparameters. The generative hyperparameter is denoted in the formulation using $\lambda_{gen}$ while the knowledge and entity-based regularization hyperparameters are denoted using $\lambda_k$, $\lambda_{\mathcal{E}}$. Each of the hyperparameters takes on values in the range of [0.1, 0.9] with increments of 0.3 and we evaluate the validation loss in each epoch during training to save the model checkpoint with the least validation loss. We experiment with three optimization configurations: i) with generative loss alone; ii) with generative loss and entity chain loss; iii) with generative loss, knowledge loss, and entity chain loss.

\begin{equation}
\mathcal{L}_{total} = \lambda_{gen}\cdot\mathcal{L}_{gen} + \lambda_{k} \cdot \mathcal{L}_{k} + \lambda_{\mathcal{E}} \cdot \mathcal{L}_{\mathcal{E}}    
\end{equation}

Each of the loss functions is based on cross-entropy criterion.

\begin{equation}
\mathcal L_{\theta} = -\frac{1}{n}\sum_{k=1}^n\mathcal P(t_k|t_{<k}, \chi; \theta)    
\end{equation}

Where $\chi$ - the input sequence (i.e., finding, or named entity chain in a finding, or a sequence of facts retrieved from the knowledge bases associated with named entities in a finding). The proposed models are trained with the objective of minimizing the aggregate loss function defined in Equation-1. All models are built and trained using PyTorch on Google Cloud NVIDIA Tesla T4 GPU.

\section{Experimental Results and Discussion}
Table-I shows the statistics of the datasets and Table-II shows the results of evaluation against the impressions (ground truth summary). Our experimental results show that jointly optimizing the task of traditional language modeling with task-specific objectives such as preserving entity-aware factual accuracy improves performance of a model. Specifically, we demonstrate this by leveraging three pre-trained abstractive summarization models and fine-tuning on our datasets using multi-objective optimization. As can been from Table-II, Precision-target, and Recall-target increase with our training objective as compared to the language modeling training objective used with the baseline models. As extensively discussed in the literature \cite{goodrich2019assessing, zhang2019optimizing}, we also argue that lexical measures (i.e., ROUGE) do not fully quantify the factual accuracy of a generated summary while a metric that measures entity-level overlap between a ground truth summary (impression) and a model-generated summary better reflects the extent to which semantics are preserved in abstractive summarization since named entities constitute significant semantics in a clinical text. A key limitation of our proposed approach is it is computationally more expensive and takes longer to train than with customary single task objective training. Another limitation we observed is that the proposed model training approach can be sensitive to hyperparameter initialization.

\begin{table}
\captionsetup{font=scriptsize}
\begin{center}
\caption{Statistics of the experimental datasets.}
\centering
\arrayrulecolor{black}
\begin{adjustbox}{width=0.45\textwidth}
\scalebox{0.75}{
\begin{tabular}
{| >{\centering\arraybackslash}m{0.90in} | >{\centering\arraybackslash}m{0.45in} | >{\centering\arraybackslash}m{0.45in} |
>{\centering\arraybackslash}m{0.35in} |
>{\centering\arraybackslash}m{0.80in} |
>{\centering\arraybackslash}m{0.75in} |
}

\hline
\textbf{Dataset} & \textbf{Train} & \textbf{Validation} & \textbf{Test} & \textbf{Avg \# tokens per Findings} & \textbf{Avg \# tokens per Impression} \\ 
\hline
Heart Failure (HF) & 4000 & 1091 & 1091 & 142 & 48 \\ 
\hline
IU X-Ray & 2200 & 593 & 593 & 33 & 12 \\
\hline
MIMIC-CXR & 40000 & 5000 & 5000 & 52 & 18 \\
\hline
\end{tabular}
}
\end{adjustbox}
\arrayrulecolor{black}
\end{center}
\vspace{-7mm}
\end{table}

\begin{table}
\centering
\captionsetup{font=scriptsize}
\begin{center}
\caption{Experimental results. Dual MOO refers to dual multi-objective optimization where only the generative loss and entity chain loss are jointly optimized during training. Triple MOO refers to modeling where the three loss functions are jointly optimized. Due to space constraints, we report average scores across the three datasets.}
\arrayrulecolor{black}
\begin{adjustbox}{width=0.5\textwidth}
\begin{tabular}{|c|c|c|c|c|c|c|}
\hline

\multirow{2}{*}{\textbf{Model}} & \multirow{2}{*}{\textbf{R-1}} & \multirow{2}{*}{\textbf{R-2}} & \multirow{2}{*}{\textbf{R-L}} & 


\multicolumn{3}{|c|}{\textbf{Entity-level Factual Accuracy}}\\

\cline{5-7}
 &  &  &  & \textbf{Precision-target} & \textbf{Recall-target} & \textbf{F1 score-target} \\ 
\hline
T5 Vanilla (Baseline) & \textbf{35.113} & \textbf{19.503} & \textbf{34.921} & 25.150 & 42.577 & 31.621 \\ 
\hline
\begin{tabular}[c]{@{}l@{}}T5 w/ named entities~\\(dual MOO) - Ours\end{tabular} & 32.628 & 18.361 & 33.827 & \textbf{29.672} & 46.581 & 36.252 \\ 
\hline
T5 w/ named entities /w facts (triple MOO) - Ours & 28.761 & 17.382 & 30.599 & 29.327 & \textbf{48.148} & \textbf{36.451} \\ 
\hline
BART Vanilla (Baseline) & \textbf{22.951} & \textbf{16.283} & \textbf{22.657} & 18.321 & 29.679 & 22.656 \\ 
\hline
BART w/ named entities (dual MOO) - Ours & 19.827 & 13.693 & 19.792 & 20.629 & 33.839 & 25.632 \\ 
\hline
\begin{tabular}[c]{@{}l@{}}BART w/ named entities /w facts~\\(triple MOO) - Ours\end{tabular} & 15.721 & 12.173 & 16.582 & \textbf{23.182} & \textbf{34.159} & \textbf{27.620} \\ 
\hline
Pegasus Vanilla (Baseline) & \textbf{28.193} & \textbf{11.387} & \textbf{28.079} & 21.739 & 28.593 & 24.699 \\ 
\hline
Pegasus w/ named entities (dual MOO) - Ours & 27.370 & 9.728 & 25.372 & 22.058 & \textbf{29.781} & \textbf{25.344} \\ 
\hline
\begin{tabular}[c]{@{}l@{}}Pegasus w/ named entities /w facts~\\(triple MOO) - Ours\end{tabular} & 24.263 & 7.836 & 22.174 & \textbf{25.661} & 25.349 & 25.504 \\
\hline
\end{tabular}
\end{adjustbox}
\arrayrulecolor{black}
\end{center}
\vspace{-5mm}
\end{table}

\section{Conclusion and Future Work}
In this study, we proposed a framework based on a transformer encoder-decoder network and transfer learning for clinical text summarization using knowledge-aware multi-objective optimization. We experimentally demonstrated that jointly optimizing generative loss, knowledge loss, and entity-based loss functions significantly improves the quality of generated summaries in terms of entity-level factual accuracy which is critical but less explored in the healthcare domain. In the future, we plan to extend the proposed multi-task learning framework for a different healthcare domain. Further, while the current study utilizes standard cross-entropy for each loss function, we plan to experiment with different loss functions including other entropy-based functions (e.g., KL-divergence) for the regularization components. In addition, while the knowledge retriever in the proposed approach is an independent unit from the summarizer, we plan to extend the proposed end-to-end training framework to include the knowledge retriever as one component of the framework.

\bibliographystyle{IEEEtran}

\bibliography{bibliography.bib}  
\end{document}